\documentclass[preprint,12pt,3p]{elsarticle}

\usepackage{amssymb}
\usepackage{graphicx}
\usepackage[table,xcdraw]{xcolor}
\usepackage{adjustbox}
\usepackage{tabularx, booktabs}
\usepackage{tablefootnote}
\usepackage{array}
\usepackage{tabularx}

\usepackage{footnote}
\usepackage{multicol}
\usepackage{supertabular}
\usepackage{fancyhdr}
\usepackage{tikz}
\usepackage[cmex10]{amsmath}
\usepackage{amsfonts}
\usepackage{algorithm}
\usepackage[noend]{algpseudocode}
\usepackage{array}
\usepackage{mdwmath}
\usepackage{eqparbox}
\usepackage[tight,footnotesize]{subfigure}
\usepackage[]{caption}
\usepackage{afterpage}
\usepackage{flafter} 
\usepackage{hyperref}

\journal{Neural Network}

\begin{document}

\begin{frontmatter}

\title{Multivariate LSTM-FCNs for Time Series Classification}

\author[label1]{Fazle Karim }
\address[label1]{Mechanical and Industrial Engineering, University of Illinois at Chicago, 900 W. Taylor St., Chicago, IL, 60607, USA}
\address[label2]{Computer Science, University of Illinois at Chicago, 900 W. Taylor St., Chicago, IL, 60607, USA}

\cortext[cor1]{Corresponding author}


\author[label2]{Somshubra Majumdar}
\author[label1]{Houshang Darabi\corref{cor1}}
 \ead{hdarabi@uic.edu}

\author[label1]{Samuel Harford}


\begin{abstract}

Over the past decade, multivariate time series classification has received great attention. We propose transforming the existing univariate time series classification models, the Long Short Term Memory Fully Convolutional Network (LSTM-FCN) and Attention LSTM-FCN (ALSTM-FCN), into a multivariate time series classification model by augmenting the fully convolutional block with a \textit{squeeze-and-excitation} block to further improve accuracy. Our proposed models outperform most state-of-the-art models while requiring minimum preprocessing. The proposed models work efficiently on various complex multivariate time series classification tasks such as activity recognition or action recognition. Furthermore, the proposed models are highly efficient at test time and small enough to deploy on memory constrained systems. 

\end{abstract}

\begin{keyword}
 Convolutional neural network \sep long short term memory \sep recurrent neural network \sep multivariate time series classification
 
\end{keyword}

\end{frontmatter}

\section{Introduction}

Time series data is used in various fields of studies, ranging from weather readings to psychological signals \cite{kadous2002temporal,sharabiani2017efficient,kehagias1997predictive, cui2015complex}. A time series is a sequence of data points in a time domain, typically in a uniform interval \cite{wang2016effective}. There is a significant increase of time series data being collected by sensors \cite{spiegel2011pattern}. A time series dataset can be univariate, where a sequence of measurements from the same variable are collected, or multivariate, where a sequence of measurements from multiple variables or sensors are collected \cite{prieto2015stacking}. Over the past decade, multivariate time series classification has received significant interest. Multivariate time series classifications are applied in healthcare \cite{kang2014bayesian}, phoneme classification \cite{graves2005framewise}, activity recognition, object recognition, and action recognition \cite{fu2015human,geurts2001pattern,pavlovic1999time, yu2015real}. In this paper, we propose two deep learning models that outperform existing algorithms.

Several time series classification algorithms have been developed over the years. Distance-based methods along with k-nearest neighbors have proven to be successful in classifying multivariate time series \cite{orsenigo2010combining}. Plenty of research indicates Dynamic Time Warping (DTW) as the best distance-based measure to use along k-NN \cite{seto2015multivariate}. 

In addition to distance-based metrics, other  algorithms are used. Typically, feature-based classification algorithms rely heavily on the features being extracted from the time series data \cite{xing2010brief}. However, feature extraction is arduous because intrinsic features of time series data are challenging to capture. For this reason, distance-based approaches are more successful in classifying multivariate time series data \cite{zheng2014time}.  Hidden State Conditional Random Field (HCRF) and Hidden Unit Logistic Model (HULM) are two successful feature-based algorithms which have led to state-of-the-art results on various benchmark datasets, ranging from online character recognition to activity recognition \cite{pei2017multivariate}. HCRF is a computationally expensive algorithm that detects latent structures of the input time series data using a chain of k-nominal latent variables. The number of parameters in the model increases linearly with the total number of latent states required \cite{quattoni2007hidden}. Further, datasets that require a large number of latent states tend to overfit the data. To overcome this, HULM proposes using H binary stochastic hidden units to model 2$^H$ latent structures of the data with only O(H) parameters. Results indicate HULM outperforming HCRF on most datasets \cite{pei2017multivariate}.

Traditional models, such as the naive logistic model (NL) and Fisher kernel learning (FKL) \cite{jaakkola2000discriminative}, show strong performance on a wide variety of time series classification problems. The NL logistic model is a linear logistic model that makes a prediction by summing the inner products between the model weights and feature vectors over time, which is followed by a softmax function \cite{pei2017multivariate}. The FKL model is effective on time series classification problems when based on Hidden Markov Models (HMM). Subsequently, the features or representation from the FKL model is used to train a linear SVM to make a final prediction. \cite{jaakkola2000discriminative,maaten2011learning}

Another common approach for multivariate time series classification is by applying dimensional reduction techniques or by concatenating all dimensions of a multivariate time series into a univariate time series. Symbolic Representation for Multivariate Time Series (SMTS) \cite{SMTS} applies a random forest on the multivariate time series to partition it into leaf nodes, each represented by a word to form a codebook. Every word is used with another random forest to classify the multivariate time series. Learned Pattern Similarity (LPS) \cite{LPS} is a similar model that extracts segments from the multivariate time series. These segments are used to train regression trees to find dependencies between them. Each node is represented by a word. Finally, these words are used with a similarity measure to classify the unknown multivariate time series. Ultra Fast Shapelets (UFS) \cite{UFS} obtains random shapelets from the multivariate time series and applies a linear SVM or a Random Forest classifier. Subsequently, UFS was enhanced by computing derivatives as features (dUFS) \cite{UFS}. The Auto-Regressive (AR) kernel \cite{ARkernel} applies an AR kernel-based distance measure to classify the multivariate time series. Auto-Regressive forests for multivariate time series modeling (mv-ARF) \cite{mvARF} uses a tree ensemble, where the trees are trained with different time lags. Most recently, WEASEL+MUSE \cite{schafer2017multivariate} builds a multivariate feature vector using a classical bag of patterns approach on each variable with various sliding window sizes to capture discrete features, words, and pairs of words. Subsequently, feature selection is used to remove non-discriminative features using a Chi-squared test. The final classification is obtained using a logistic classifier on the final feature vector.   

Deep learning has also yielded promising results for multivariate time series classification. In 2014, \textit{Yi et al.} propose using Multi-Channel Deep Convolutional Neural Network (MC-DCNN) for multivariate time series classification. MC-DCNN takes input from each variable to detect latent features. The latent features from each channel are fed into an MLP to perform classification \cite{zheng2014time}. 

This paper proposes two deep learning models for multivariate time series classification. These proposed models require minimal preprocessing and are tested on 35 datasets, obtaining strong performances in most of them. Performance is the classification accuracy of a model on a particular dataset. The rest of the paper is ordered as follows. Background works are discussed in Section \ref{Background Works}. We present the architecture of the two proposed models in Section \ref{LSTMFCN}. In Section \ref{Experiments}, we discuss the dataset, evaluate the models on them, present our results and analyze our findings. In Section \ref{conclusion}, we draw our conclusion.

\section{Background Works}
\label{Background Works}
\subsection{Recurrent Neural Networks}
\def\x{{\mathbf x}}
\def\L{{\cal L}}

Recurrent Neural Networks (RNN) are a form of neural networks that display temporal behavior through the direct connections between individual layers.  \textit{Pascanu et al.} \cite{pascanu2013construct} implement RNN to maintain a hidden vector $\mathbf h$ that is updated at time step $t$,
\begin{equation}
	\mathbf h_t = \tanh(\mathbf W\mathbf h_{t-1} + \mathbf I\mathbf  \x_t),
\end{equation}
where the hyperbolic tangent function is represented by $tanh$, the input vector at time step $t$ is denoted as $\mathbf x_t$, the recurrent weight matrix is denoted by $\mathbf W$ and the projection matrix is signified by $\mathbf I$. A prediction, $\mathbf y_t$ can be made using a hidden state,  $\mathbf h$, and a weight matrix, $\mathbf W$,
\begin{equation}
	\mathbf y_t = \text{softmax}(\mathbf W\mathbf h_{t-1}).
\end{equation}
The softmax function normalizes the output predictions of the model to be a valid probability distribution and the logistic sigmoid function is declared as $\sigma$. RNNs can be stacked to create deeper networks by using the hidden state,  $\mathbf h^{l-1}$ of a RNN layer $l{-1}$ as an input to the hidden state, $\mathbf h^l$ of another RNN layer $l$, 
\begin{equation}
	\mathbf h_t^{l} = \sigma(\mathbf W\mathbf h_{t-1}^{l} + \mathbf I\mathbf h_t^{l-1}).
\end{equation}

\subsection{Long Short-Term Memory RNNs}
\def\x{{\mathbf x}}

A major issue with RNNs is that they often have to face the issue of vanishing gradients. Long short-term memory (LSTM) RNNs address this problem by integrating gating functions into their state dynamics \cite{hochreiter1997long}. An LSTM maintains a hidden vector, $\mathbf h$, and a memory vector, $\mathbf m$, which control state updates and outputs at each time step, respectively.  The computation at each time step is depicted by \textit{Graves et al.} \cite{graves2012supervised} as the following:
\begin{equation}
	\begin{split}
		& \mathbf g^u = \sigma(\mathbf W^u\mathbf h_{t-1}  + \mathbf I^u\x_t ) \\
		& \mathbf g^f = \sigma(\mathbf W^f\mathbf h_{t-1} + \mathbf I^f\x_t) \\
		& \mathbf g^o = \sigma(\mathbf W^o\mathbf h_{t-1} + \mathbf I^o\x_t) \\
		& \mathbf g^c = \tanh(\mathbf W^c\mathbf h_{t-1} + \mathbf I^c\x_t) \\
		& \mathbf m_t = \mathbf g^f \odot \mathbf m_{t-1}\mathbf \ + \  \mathbf g^u \odot 
		\mathbf g^c \\
		& \mathbf h_t = \tanh(\mathbf g^o \odot \mathbf m_t) 
	\end{split}
\end{equation}
where \textbf{$\mathbf g^u$}, \textbf{$\mathbf g^f$}, \textbf{$\mathbf g^o$}, \textbf{$\mathbf g^c$} are the activation vectors of the input, forget, output and cell state gates respectively, $h_t$ is the hidden state vector of the LSTM unit, the logistic sigmoid function is defined by $\sigma$, the elementwise multiplication is represented by $\odot$. The recurrent weight matrices are depicted using the notation $\mathbf W^u, \mathbf W^f, \mathbf W^o, \mathbf W^c$ and the projection matrices are portrayed by $\mathbf I^u, \mathbf I^f, \mathbf I^o, \mathbf I^c$.

LSTMs can learn temporal dependencies. However, long-term dependencies of long sequence are challenging to learn using LSTMs. \textit{Bahdanau et al.} \cite{bahdanau2014neural} proposed using an attention mechanism to learn these long-term dependencies. 

\subsection{Attention Mechanism}
An attention mechanism conditions a context vector $V$ on the target sequence $y$. This method is commonly used in neural translation of texts. \textit{Bahdanau et al.}\cite{bahdanau2014neural} argues the context vector $v_i$ depends on a sequence of \textit{annotations} $(b_1, ..., b_{T_{x}})$, of length $T_x$ which is the maximum length of the input sequence $x$, where an encoder maps the input sequence. Each annotation, $b_i$, comprises information on the whole input sequence, while focusing on regions surrounding the $i$-th word of the input sequence. The weighted sum of each annotation, $b_i$, is used to compute the context vector as follows:
\begin{equation}
    v_i = \sum_{j=1}^{T_x} \alpha_{ij}b_j.
\end{equation}

The weight, $\alpha_{ij}$, of each annotation is calculated through : 
\begin{equation}
    \alpha_{ij} = \frac{exp(e_{ij})}{\sum_{k=1}^{T_x} exp(e_{ik})},
\end{equation}

where the energy of alignment, $e_{ij}$, is given by $a(\nu_{i-1}, b_j)$, which measures how well the input position, $j$, and the output at position, $i$, match using the RNN hidden state, $\nu_{i-1}$, and the $j$-th annotation, $b_j$, of the input sequence. \textit{Bahdanau et al.}\cite{bahdanau2014neural} uses a feedforward neural network to parameterize the alignment model, $a$. The feedforward neural network is trained jointly with all other components of the model. In addition, the alignment model calculates a soft alignment that can backpropagate the gradient of the cost function. The gradient of the cost function trains the alignment model and the translation model simultaneously \cite{bahdanau2014neural}.

\subsection{Squeeze-and-Excitation Block}
\textit{Hu  et al} \cite{hu2017squeeze} propose a  \textit{squeeze-and-excitation} block that acts as a computational unit for any transformation $\textbf{F}_{tr} : \textbf{X} \rightarrow \textbf{U}, \textbf{X} \in \mathbb{R}^{W' \times H' \times C'}, \textbf{U} \in \mathbb{R}^{W \times H \times C}$. The outputs of $\textbf{F}_{tr}$ are represented as U = $[\textbf{u}_1, \textbf{u}_2, \cdots, \textbf{u}_C]$ where 
\begin{equation}
    \textbf{u}_c = \textbf{v}_c * \textbf{X} = \sum_{s=1}^{C'}{\textbf{v}_c^s*\textbf{x}^s}
\end{equation}
The convolution operation is depicted by *, and the 2D spatial kernel is depicted by $\mathbf v_c^s$. The single channel of v$_c$ acts on the corresponding channel of X. \textit{Hu  et al.} \cite{hu2017squeeze} models the channel interdependencies to adjust the filter responses in two steps, $squeeze$ and $excitation$. 

The $squeeze$ operation exploits the contextual information outside the local receptive field by using a global average pool to generate channel-wise statistics. The transformation output, \textbf{U}, is shrunk through spatial dimensions, $W \times  H$, to compute the channel-wise statistics, z $\in \mathbb{R}^C$. The \textit{c}-th element of \textbf{z} is calculated by computing $\mathbf F_{sq}(\mathbf u_c)$, which is the channel-wise global average over the spatial dimensions $W \times H$, defined as:
\begin{equation}
    \textbf{z}_c = \textbf{F}_{sq}(\textbf{u}_c)=\frac{1}{W \times H} \sum_{i=1}^{W} \sum_{j=1}^{H} u_c(i,j)
\end{equation}
For temporal sequence data, the transformation output, \textbf{U}, is shrunk through the temporal dimension $T$ to compute the channel-wise statistics, z $\in \mathbb{R}^C$. The \textit{c}-th element of  \textbf{z} is then calculated by computing $\mathbf F_{sq}(\mathbf u_c)$, which is the channel-wise global average over the temporal dimension $T$, defined as:
\begin{equation}
    \textbf{z}_c = \textbf{F}_{sq}(\textbf{u}_c)=\frac{1}{T} \sum_{t=1}^{T} u_c(t)
\end{equation}

The aggregated information obtained from the $squeeze$ operation is followed by an $excite$ operation, whose objective is to capture the channel-wise dependencies. To achieve this, a simple gating mechanism is applied with a sigmoid activation, as follows:
\begin{equation}
    \textbf{s} = \textbf{F}_{ex} (\textbf{z,W}) = \sigma (g(\textbf{z,W})) = \sigma (\textbf{W}_2 \delta (\textbf{W}_1 z)),
\end{equation}
where $\mathbf F_{ex}$ is parameterized as a neural network, $\sigma$ is the Sigmoid activation function, $\delta$ is the ReLU activation function, $\textbf{W}_1 \in \mathbb{R}^{\frac{C}{r}\times C}$ and $\textbf{W}_2 \in \mathbb{R}^{\frac{C}{r}\times C}$ are learnable parameters of $\mathbf F_{ex}$ and \textit{r} is the reduction ratio. $\textbf{W}_1$ and $\textbf{W}_2$  are used to limit model complexity and aid with generalization. $\textbf{W}_1$ are the parameters of a dimensionality-reduction layer and $\textbf{W}_2$ are the parameters of a dimensionality-increasing layer. 

Finally, the output of the block is rescaled as follows:
\begin{equation}
\widetilde{\textbf{x}}_c = \textbf{F}_{scale} (\textbf{u}_c, \textbf{s}_c) = \textbf{s}_c \cdot \textbf{u}_c,  
\end{equation}
where $\widetilde{\textbf{X}} = [\widetilde{\textbf{x}}_1, \widetilde{\textbf{x}}_2, ... , \widetilde{\textbf{x}}_C]$ and $\textbf{F}_{scale}(\textbf{u}_c, \textbf{s}_c)$ denotes the channel-wise multiplication between the feature map $\textbf{u}_c \in \mathbb{R}^{T}$ and the scale $s_c$.

\section{Multivariate LSTM Fully Convolutional Network}
\label{LSTMFCN}
\subsection{Network Architecture}

Long Short Term Memory Fully Convolutional Network (LSTM-FCN) and Attention LSTM-FCN (ALSTM-FCN) have been successful in classifying univariate time series \cite{karim2017lstm}. However, they have never been applied to on a multivariate time series classification problem. The models we propose, Multivariate LSTM-FCN (MLSTM-FCN) and Multivariate Attention LSTM-FCN (MALSTM-FCN), converts their respective univariate models into multivariate variants. We extend the \textit{squeeze-and-excite} block to the case of 1D sequence models and augment the fully convolutional blocks of the LSTM-FCN and ALSTM-FCN models to enhance classification accuracy.

As the datasets now consist of multivariate time series, we can define a time series dataset as a tensor of shape (\textit{N, Q, M}), where \textit{N} is the number of samples in the dataset, Q is the maximum number of time steps amongst all variables and M is the number of variables processed per time step. Therefore a univariate time series dataset is a special case of the above definition, where \textit{M} is 1. The alteration required to the input of the LSTM-FCN and ALSTM-FCN models is to accept \textit{M} inputs per time step, rather than a single input per time step.

\begin{figure*}[htpb]
\center
\includegraphics[width=0.75\linewidth]{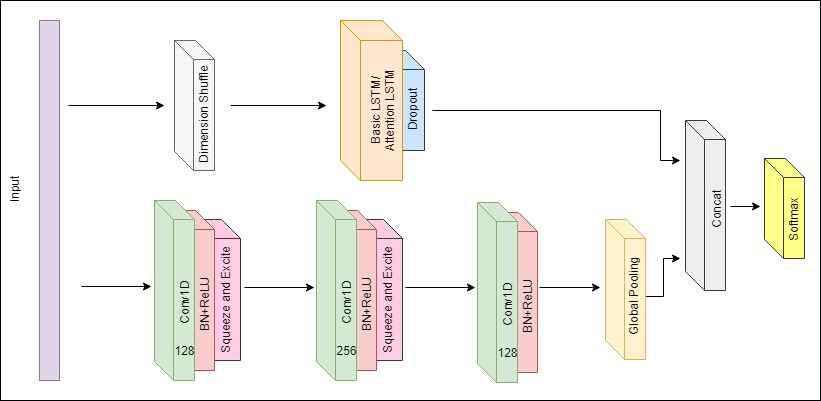}

\center
\caption{The MLSTM-FCN architecture. LSTM cells can be replaced by Attention LSTM cells to construct the MALSTM-FCN architecture.}
\label{fig:arch}

\end{figure*}

\begin{figure*}[htpb]
\center
\includegraphics[width=0.85\linewidth]{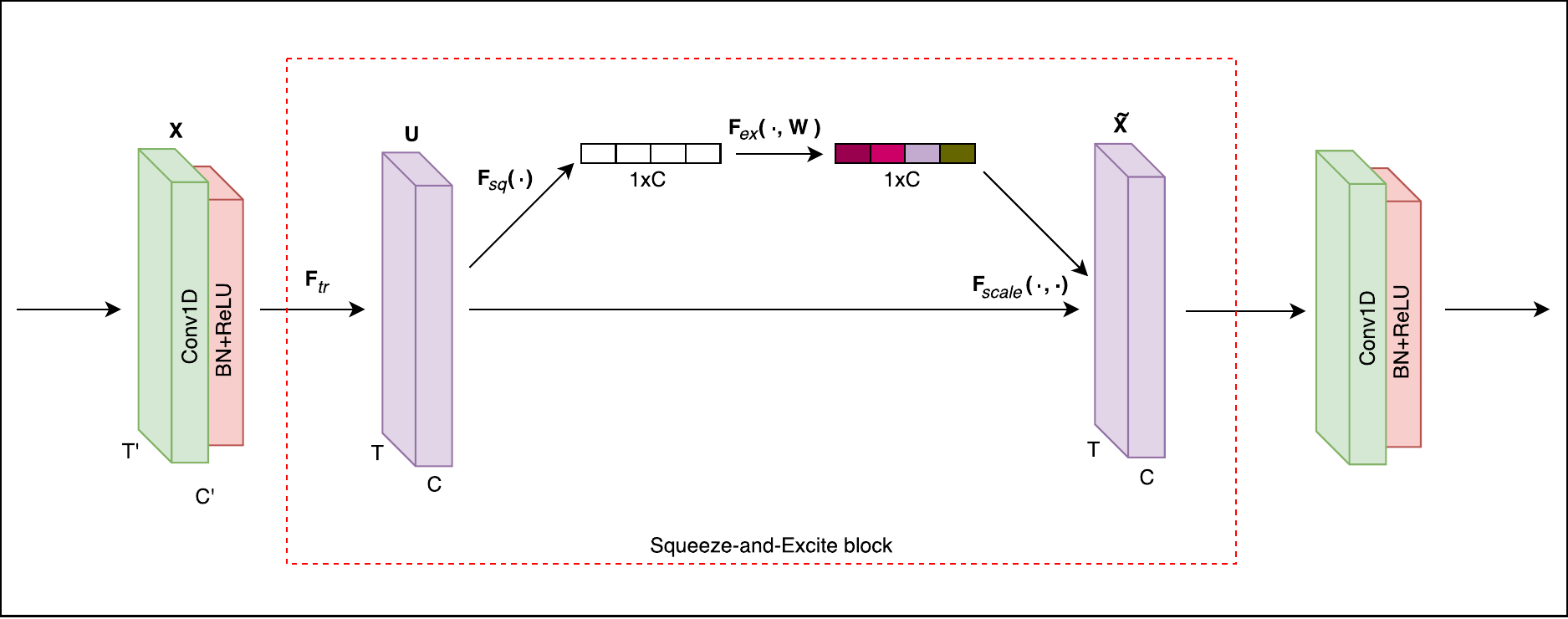}

\center
\caption{The computation of the temporal \textit{squeeze-and-excite} block.}
\label{fig:squeeze}

\end{figure*}

Similar to LSTM-FCN and ALSTM-FCN, the proposed models comprise a fully convolutional block and a LSTM block, as depicted in Fig. \ref{fig:arch}. The fully convolutional block contains three temporal convolutional blocks, used as a feature extractor, which is replicated from the original fully convolutional block by \textit{Wang et al} \cite{wang2017time}. The convolutional blocks contain a convolutional layer with a number of filters (128, 256, and 128) and a kernel size of 8, 5, and 3 respectively. Each convolutional layer is succeeded by batch normalization, with a momentum of 0.99 and epsilon of 0.001. The batch normalization layer is succeeded by the ReLU activation function. In addition, the first two convolutional blocks conclude with a \textit{squeeze-and-excite} block, which sets the proposed model apart from LSTM-FCN and ALSTM-FCN. Fig. \ref{fig:squeeze} summarizes the process of how the \textit{squeeze-and-excite} block is computed in our architecture.  For all \textit{squeeze and excitation} blocks, we set the reduction ratio \textit{r} to 16. The final temporal convolutional block is followed by a global average pooling layer. 

The \textit{squeeze-and-excite} block is an addition to the FCN block which adaptively recalibrates the input feature maps. Due to the reduction ratio \textit{r} set to 16, the number of parameters required to learn these self-attention maps is reduced such that the overall model size increases by just 3-10 \%. This can be computed as below:
 
\begin{equation*}
    P = \frac{2}{r} \sum_{s=1}^S R_s \cdot G_s^2
\end{equation*}
 
where \textit{P} is the total number of additional parameters, \textit{r} denotes the reduction ratio, \textit{S} denotes the number of stages (each stage refers to the collection of blocks operating on feature maps of a common spatial dimension), $G_s$ denotes the number of output feature maps for stage $s$ and $R_s$ denotes the repeated block number for stage $s$. Since the FCN blocks are kept consistent for all models, we can directly compute the additional number of parameters as $P = \frac{2}{16} * (128^2 + 256^2) = 10240$ for all models. \textit{Squeeze} and \textit{excitation} is essential to the enhanced performance on multivariate datasets, as not all feature maps may impact the subsequent layers to the same degree. This adaptive recalibration of the feature maps can be considered as a form of learned self-attention on the output feature maps of prior layers. This adaptive rescaling of the filter maps is of utmost importance to the improved performance of the MLSTM-FCN model compared to LSTM-FCN, as it incorporates learned self-attention to the inter-correlations between multiple variables at each time step, which was inadequate with the LSTM-FCN.

In addition, the multivariate time series input is passed through a dimension shuffle layer (explained more in Section \ref{net_input}), followed by the LSTM block. The LSTM block is identical to the block from the LSTM-FCN or ALSTM-FCN models \cite{karim2017lstm}, comprising either an LSTM layer or an Attention LSTM layer, which is followed by a dropout layer. 

\subsection{Network Input}
\label{net_input}
Depending on the dataset, the input to the fully convolutional block and LSTM block vary. The input to the fully convolutional block is a multivariate variate time series with \textit{Q} time steps having \textit{M} distinct variables per time step. If there is a time series with \textit{M} variables and \textit{Q} time steps, the fully convolutional block will receive the data as such. Variables are defined as the channels of interconnected data streams.  

In addition, the input to the LSTM can vary depending on the application of dimension shuffle. The dimension shuffle transposes the temporal dimension of the input data. If the dimension shuffle operation is not applied to the LSTM path, the LSTM will require \textit{Q} time steps to process \textit{M} variables at each time step. However, if the dimension shuffle is applied, the LSTM will require \textit{M} time steps to process \textit{Q} variables per time step. In other words, the dimension shuffle improves the efficiency of the model when the number of variables \textit{M} is less than the number of time steps \textit{Q}.

After the dimension shuffle, at each time step $t$, where $1 \leq t \leq M$, \textit{M} being the number of variables, the input provides the LSTM the entire history of that variable (data of that variable over all Q time steps). Thus, the LSTM is given the global temporal information of each variable at once. As a result, the dimension shuffle operation reduces the computation time of training and inference without losing accuracy for time series classification problems. An ablation test is performed to show the performance of a model with the dimension shuffle operation is statistically the same as a model without using it (further discussed in Section \ref{ablation}).

The proposed models take a total of 13 hours to process the MLSTM-FCN and a total of 18 hours to process the MALSTM-FCN on a single GTX 1080 Ti GPU. While the time required to train these models is significant, one can note their inference time is comparable with other standard models.

\section{Experiments}
\label{Experiments}
MLSTM-FCN and MALSTM-FCN have been tested on 35 datasets, in Section \ref{dataset_section}. The optimal number of LSTM cells for each dataset was found via grid search over 3 distinct choices - 8, 64 or 128, and all other hyper parameters are kept constant. The FCN block is comprised of 3 blocks of 128-256-128 filters for all models, with kernel sizes of 8, 5, and 3 respectively, comparable with the original models proposed by \textit{Wang et al} \cite{wang2017time}. Additionally, the first two FCN blocks are succeded by the \textit{squeeze-and-excitation} block. We consistently chose 16 as the reduction ratio \textit{r} for all \textit{squeeze-and-excitation} blocks, as suggested by \textit{Hu et al} \cite{hu2017squeeze}. During the training phase, we set the total number of training epochs to 250 unless explicitly stated and the dropout rate is set to 80\% to mitigate overfitting. Each of the proposed models is trained using a batch size of 128. The convolution kernels are initialized by the \textit{Uniform He} initialization scheme proposed by \textit{He et al} \cite{he2015delving}, which samples from the uniform distribution $U \in \left (-\sqrt{\frac{6}{d}}, \sqrt{\frac{6}{d}} \right )$, where \textit{d} is the number of input units to the weight tensor.  For datasets with class imbalance, a class weighing scheme inspired by \textit{King et al.} is utilized \cite{king2001logistic}, weighing the contribution of each class $C_i$ ($1 \leq i \leq C$) to the loss by the factor $Gw_i = \frac{N}{C \times N_{C_i}}$, where $Gw_i$ is the loss scaling weight for the \textit{i}-th class, \textit{N} is the number of samples in the dataset, \textit{C} is the number of classes and $N_{C_i}$ is a the number of samples which belong to class $C_i$. 

We use the Adam optimizer \cite{kingma2014adam}, with an initial learning rate set to 1e-3 and the final learning rate set to 1e-4 to train all models. In addition, after every 100 epochs, the learning rate is reduced by a factor of $1/{\sqrt[3]{2}}$. The datasets were normalized and preprocessed to have zero mean and unit variance. We append variable length time series with zeros afterwards to obtain a time series dataset with a constant length \textit{Q}, where \textit{Q} is the maximum length of the time series. Mean-standard deviation normalization and zero padding are the only preprocessing steps performed for all models. We compute the mean and standard deviation of the training dataset and apply these values to normalize both the train and test datasets. We use the Keras \cite{chollet2015keras} library with the Tensorflow backend \cite{tensorflow2015-whitepaper} to train the proposed models.

\subsection{Evaluation Metrics}


In this paper, various models, including the proposed models, are evaluated using accuracy, arithmetic rank, geometric rank, the Wilcoxon signed-rank test, and mean per class error. The arithmetic and geometric rank are the arithmetic and geometric mean of the ranks,

\begin{align*}
\text{Arithmetic Mean}_{K} &= \frac{\sum_{K}^{\textbf{N}} \textit{rank}_{K}}{\textbf{N}} \\
\text{Geometric Mean}_{K} &= \frac{\prod_{K}^{\textbf{N}} \textit{rank}_K}{\textbf{N}},
\end{align*}
where $K$ is the dataset and $\textbf{N}$ is the number of datasets. 

The Wilcoxon signed-rank test is a non-parametric statistical test that hypothesizes the median of the rank between the compared models is the same. The alternative hypothesis of the Wilcoxon signed-rank test is that the median of the rank between the compared models is not the same. Finally, the mean per class error is the average error of each class for all the datasets,
\begin{align*}
PCE_k &=\frac{1-accuracy}{\textit{number of unique classes}} \\
MPCE &=\frac{1}{{\textbf{N}}} \sum{PCE_K}.
\end{align*}

\subsection{Datasets}
\label{dataset_section}
 
\def\tabularxcolumn#1{m{#1}}
\setlength\arrayrulewidth{1pt}

\newcolumntype{b}{>{\centering \arraybackslash\hsize=1.5 \hsize}X}
\newcolumntype{s}{>{\centering \arraybackslash \hsize=.5 \hsize}X}
\newcommand{\heading}[1]{\multicolumn{1}{c}{#1}}

\begin{table}[]
\centering
\caption{Properties of all datasets. The yellow cells are datasets used by \textit{Pei et al.} \cite{pei2017multivariate}, the purple cells are datasets used by \textit{Sch\"{a}fer and Leser} \cite{schafer2017multivariate}, and the blue cells are datasets from the UCI repository \cite{Lichman:2013}.}
\label{tab:sum_data}
\begin{adjustbox}{width=1 \textwidth}

\begin{tabularx}{1.5 \textwidth}{|b|s|s|s|b|b|c|}

    \hline
    \textbf{Dataset} & \textbf{Num. of Classes} & \textbf{Num. of Variables} & \textbf{Max Training Length} & \textbf{Task} & \textbf{Train-Test Split} & \textbf{Source}\\
    \hline
    \rowcolor[rgb]{ 1,  .98,  .81} \textbf{OHC} & 20    & 30    & 173   & Handwriting Classification & 10-fold & \cite{williams2008modelling}\\
    \hline
    \rowcolor[rgb]{ 1,  .98,  .81} \textbf{Arabic Voice} & 88    & 39    & 91    & Speaker Recognition & 75-25 split & \cite{hammami2010improved} \\
    \hline
    \rowcolor[rgb]{ 1,  .98,  .81} \textbf{Cohn-Kanade AU-coded Expression (CK+)} & 7     & 136   & 71    & Facial Expression Classification & 10-fold & \cite{van2012action} \\
    \hline
    \rowcolor[rgb]{ 1,  .98,  .81} \textbf{MSR Action} & 20    & 570   & 100   & Action Recognition & 5 ppl in train; rest in test & \cite{li2010action} \\
    \hline
    \rowcolor[rgb]{ 1,  .98,  .81} \textbf{MSR Activity} & 16    & 570   & 337   & Activity Recognition & 5 ppl in train; rest in test & \cite{wang2012mining} \\
    \hline
    \rowcolor[rgb]{ .87,  .81,  1} \textbf{ArabicDigits} & 10    & 13    & 93    & Digit Recognition & 75-25 split & \cite{Lichman:2013} \\
    \hline
    \rowcolor[rgb]{ .87,  .81,  1} \textbf{AUSLAN} & 95    & 22    & 96    & Sign Language Recognition & 44-56 split & \cite{Lichman:2013} \\
    \hline
    \rowcolor[rgb]{ .87,  .81,  1} \textbf{CharacterTrajectories} & 20    & 3     & 205   & Handwriting Classification & 10-90 split & \cite{Lichman:2013} \\
    \hline
    \rowcolor[rgb]{ .87,  .81,  1} \textbf{CMU\_MOCAP\_S16} & 2     & 62    & 534   & Action Recognition & 50-50 split & \cite{cmu} \\
    \hline
    \rowcolor[rgb]{ .87,  .81,  1} \textbf{DigitShape} & 4     & 2     & 97    & Action Recognition & 60-40 split & \cite{subakan2014probabilistic} \\
    \hline
    \rowcolor[rgb]{ .87,  .81,  1} \textbf{ECG} & 2     & 2     & 147   & ECG Classification & 50-50 split & \cite{bobski_world} \\
    \hline
    \rowcolor[rgb]{ .87,  .81,  1} \textbf{JapaneseVowels} & 9     & 12    & 26    & Speech Recognition & 42-58 split & \cite{Lichman:2013} \\
    \hline
    \rowcolor[rgb]{ .87,  .81,  1} \textbf{KickvsPunch} & 2     & 62    & 761   & Action Recognition & 62-38 split & \cite{cmu} \\
    \hline
    \rowcolor[rgb]{ .87,  .81,  1} \textbf{LIBRAS} & 15    & 2     & 45    & Sign Language Recognition & 38-62 split & \cite{Lichman:2013} \\
    \hline
    \rowcolor[rgb]{ .87,  .81,  1} \textbf{LP1} & 4     & 6     & 15    & Robot Failure Recogntion & 43-57 split & \cite{Lichman:2013} \\
    \hline
    \rowcolor[rgb]{ .87,  .81,  1} \textbf{LP2} & 5     & 6     & 15    & Robot Failure Recogntion & 36-64 split & \cite{Lichman:2013} \\
    \hline
    \rowcolor[rgb]{ .87,  .81,  1} \textbf{LP3} & 4     & 6     & 15    & Robot Failure Recogntion & 36-64 split & \cite{Lichman:2013} \\
    \hline
    \rowcolor[rgb]{ .87,  .81,  1} \textbf{LP4} & 3     & 6     & 15    & Robot Failure Recogntion & 36-64 split & \cite{Lichman:2013} \\
    \hline
    \rowcolor[rgb]{ .87,  .81,  1} \textbf{LP5} & 5     & 6     & 15    & Robot Failure Recogntion & 39-61 split & \cite{Lichman:2013} \\
    \hline
    \rowcolor[rgb]{ .87,  .81,  1} \textbf{NetFlow} & 2     & 4     & 994   & Action Recognition & 60-40 split & \cite{subakan2014probabilistic} \\
    \hline
    \rowcolor[rgb]{ .87,  .81,  1} \textbf{PenDigits} & 10    & 2     & 8     & Digit Recognition & 2-98 split & \cite{Lichman:2013} \\
    \hline
    \rowcolor[rgb]{ .87,  .81,  1} \textbf{Shapes} & 3     & 2     & 97    & Action Recognition & 60-40 split & \cite{subakan2014probabilistic} \\
    \hline
    \rowcolor[rgb]{ .87,  .81,  1} \textbf{Uwave} & 8     & 3     & 315   & Gesture Recognition & 20-80 split & \cite{Lichman:2013} \\
    \hline
    \rowcolor[rgb]{ .87,  .81,  1} \textbf{Wafer} & 2     & 6     & 198   & Manufacturing Classification & 25-75 split & \cite{bobski_world} \\
    \hline
    \rowcolor[rgb]{ .87,  .81,  1} \textbf{WalkVsRun} & 2     & 62    & 1918  & Action Recognition & 64-36 split & \cite{cmu} \\
    \hline
    \rowcolor[rgb]{ .81,  .87,  1} \textbf{AREM} & 7     & 7     & 480   & Activity Recognition & 50-50 split & \cite{Lichman:2013} \\
    \hline
    \rowcolor[rgb]{ .81,  .87,  1} \textbf{HAR} & 6     & 9     & 128   & Activity Recognition & 71-29 split & \cite{Lichman:2013} \\
    \hline
    \rowcolor[rgb]{ .81,  .87,  1} \textbf{Daily Sport} & 19    & 45    & 125   & Activity Recognition & 50-50 split & \cite{Lichman:2013} \\
    \hline
    \rowcolor[rgb]{ .81,  .87,  1} \textbf{Gesture Phase} & 5     & 18    & 214   & Gesture Recognition & 50-50 split & \cite{Lichman:2013} \\
    \hline
    \rowcolor[rgb]{ .81,  .87,  1} \textbf{EEG} & 2     & 13    & 117   & EEG Classification & 50-50 split & \cite{Lichman:2013} \\
    \hline
    \rowcolor[rgb]{ .81,  .87,  1} \textbf{EEG2} & 2     & 64    & 256   & EEG Classification & 20-80 split & \cite{Lichman:2013} \\
    \hline
    \rowcolor[rgb]{ .81,  .87,  1} \textbf{HT Sensor} & 3     & 11    & 5396  & Food Classification & 50-50 split & \cite{Lichman:2013} \\
    \hline
    \rowcolor[rgb]{ .81,  .87,  1} \textbf{Movement AAL} & 2     & 4     & 119   & Movement Classification & 50-50 split & \cite{Lichman:2013} \\
    \hline
    \rowcolor[rgb]{ .81,  .87,  1} \textbf{Occupancy} & 2     & 5     & 3758  & Occupancy Classification & 35-65 split & \cite{Lichman:2013} \\
    \hline
    \rowcolor[rgb]{ .81,  .87,  1} \textbf{Ozone} & 2     & 72    & 291   & Weather Classification & 50-50 split & \cite{Lichman:2013} \\
    \hline

\end{tabularx}
\end{adjustbox}

\end{table}
A total of 35 datasets are used to test the proposed models. Five of the 35 datasets are benchmark datasets used by \textit{Pei et al.}\cite{pei2017multivariate}, where the training and testing sets are provided online. In addition, we test the proposed models on 20 benchmark datasets, most recently utilized by \textit{Sch\"{a}fer and Leser} \cite{schafer2017multivariate}. These 20 datasets are trained on the same training and testing datasets as \textit{Sch\"{a}fer and Leser} \cite{schafer2017multivariate}. These benchmark datasets are from various fields. Some datasets encompass the domains of medical care, speech recognition and motion recognition. Further details of each dataset are depicted in Table \ref{tab:sum_data}. The max training length in Table \ref{tab:sum_data} is the maximum number of time steps for the entire sequence. The remaining 10 datasets of various classification tasks were obtained from the UCI repository \cite{Lichman:2013}. ``HAR'', ``EEG2'', and the ``Occupancy'' datasets have predefined training and testing sets. All the remaining datasets are partitioned into training and testing sets with a split ratio of 50:50. Each of the datasets is normalized to have zero mean and unit standard deviation. Furthermore, the datasets are padded with zeros, such that each time series length is equivalent to the maximum length of all variables in the training dataset. The dataset is summarized in Table \ref{tab:sum_data}.

\subsection{Results}
 \newcolumntype{C}{>{\centering\arraybackslash}X}
 \newcommand\mcxl[1]{\multicolumn{1}{|C|}{\bfseries #1}}
 \newcommand\mcx[1]{\multicolumn{1}{C }{\bfseries #1}}
 \newcolumntype{L}{@{}>{\iffalse}l<{\fi}}



 \begin{table*}[htpb]
 \centering
 \caption{Performance comparison of proposed models with the rest on benchmark datasets. Green cells denote model with best performance. Red font indicates models that have a strided convolution prior to the LSTM block.}
\label{tab:perf_tab}
\begin{adjustbox}{width=1 \linewidth}

 \begin{tabularx}{1.4 \textwidth}{|C|C|C|C|C|C|}
    \hline
    \textbf{Datasets} & \textbf{LSTM-FCN} & \textbf{MLSTM-FCN} & \textbf{ALSTM-FCN} & \textbf{MALSTM-FCN} & {\textbf{SOTA}} \\
    \hline
   {\textbf{Action 3d}} & \cellcolor[rgb]{ .663,  .816,  .557}71.72 & \cellcolor[rgb]{ .663,  .816,  .557}75.42 & \cellcolor[rgb]{ .663,  .816,  .557}72.73 & \cellcolor[rgb]{ .663,  .816,  .557}\textcolor[rgb]{ 1,  0,  0}{74.74} & 70.71 [DTW] \\
    \hline
   {\textbf{Activity}} & 53.13 & 61.88 & 55.63 & \textcolor[rgb]{ 1,  0,  0}{58.75} & 66.25 [DTW] \\
    \hline
    \textbf{ArabicDigits} & \cellcolor[rgb]{ .663,  .816,  .557}100.00 & \cellcolor[rgb]{ .663,  .816,  .557}100.00 & \cellcolor[rgb]{ .663,  .816,  .557}99.00 & \cellcolor[rgb]{ .663,  .816,  .557}99.00 & 99.00 \cite{schafer2017multivariate} \\
    \hline
   {\textbf{Arabic-Voice}} & \cellcolor[rgb]{ .663,  .816,  .557}98.00 & \cellcolor[rgb]{ .663,  .816,  .557}98.00 & \cellcolor[rgb]{ .663,  .816,  .557}98.55 & \cellcolor[rgb]{ .663,  .816,  .557}98.27 & 94.55 \cite{pei2017multivariate} \\
    \hline
    \textbf{AREM} & \cellcolor[rgb]{ .663,  .816,  .557}76.92 & \cellcolor[rgb]{ .663,  .816,  .557}84.62 & \cellcolor[rgb]{ .663,  .816,  .557}76.92 & \cellcolor[rgb]{ .663,  .816,  .557}84.62 & 76.92 [DTW] \\
    \hline
    \textbf{AUSLAN} & 97.00 & 97.00 & 96.00 & 96.00 & 98.00 \cite{schafer2017multivariate} \\
    \hline
    \textbf{CharacterTrajectories} & \cellcolor[rgb]{ .663,  .816,  .557}99.00 & \cellcolor[rgb]{ .663,  .816,  .557}100.00 & \cellcolor[rgb]{ .663,  .816,  .557}99.00 & \cellcolor[rgb]{ .663,  .816,  .557}100.00 & 99.00 \cite{schafer2017multivariate} \\
    \hline
   {\textbf{CK+}} & \cellcolor[rgb]{ .663,  .816,  .557}96.07 & \cellcolor[rgb]{ .663,  .816,  .557}96.43 & \cellcolor[rgb]{ .663,  .816,  .557}97.10 & \cellcolor[rgb]{ .663,  .816,  .557}97.50 & 93.56 \cite{pei2017multivariate} \\
    \hline
    \textbf{CMUsubject16} & \cellcolor[rgb]{ .663,  .816,  .557}100.00 & \cellcolor[rgb]{ .663,  .816,  .557}100.00 & \cellcolor[rgb]{ .663,  .816,  .557}100.00 & \cellcolor[rgb]{ .663,  .816,  .557}100.00 & 100.00 \cite{mvARF} \\
    \hline
    \textbf{Daily Sport} & \cellcolor[rgb]{ .663,  .816,  .557}99.65 & \cellcolor[rgb]{ .663,  .816,  .557}99.65 & \cellcolor[rgb]{ .663,  .816,  .557}99.63 & \cellcolor[rgb]{ .663,  .816,  .557}99.72 & 98.42 [DTW] \\
    \hline
    \textbf{DigitShapes} & \cellcolor[rgb]{ .663,  .816,  .557}100.00 & \cellcolor[rgb]{ .663,  .816,  .557}100.00 & \cellcolor[rgb]{ .663,  .816,  .557}100.00 & \cellcolor[rgb]{ .663,  .816,  .557}100.00 & 100.00 \cite{mvARF} \\
    \hline
    \textbf{ECG} & 85.00 & 86.00 & 86.00 & 86.00 & 93.00 \cite{schafer2017multivariate} \\
    \hline
    \textbf{EEG} & 60.94 & \cellcolor[rgb]{ .663,  .816,  .557}65.63 & \cellcolor[rgb]{ .663,  .816,  .557}64.06 & \cellcolor[rgb]{ .663,  .816,  .557}64.07 & 62.50 [RF] \\
    \hline
    \textbf{EEG2} & \cellcolor[rgb]{ .663,  .816,  .557}90.67 & \cellcolor[rgb]{ .663,  .816,  .557}91.00 & \cellcolor[rgb]{ .663,  .816,  .557}90.67 & \cellcolor[rgb]{ .663,  .816,  .557}91.33 & 77.50 [RF] \\
    \hline
    \textbf{Gesture Phase} & \cellcolor[rgb]{ .663,  .816,  .557}50.51 & \cellcolor[rgb]{ .663,  .816,  .557}53.53 & \cellcolor[rgb]{ .663,  .816,  .557}52.53 & \cellcolor[rgb]{ .663,  .816,  .557}53.05 & 40.91 [DTW] \\
    \hline
    \textbf{HAR} & \cellcolor[rgb]{ .663,  .816,  .557}96.00 & \cellcolor[rgb]{ .663,  .816,  .557}96.71 & \cellcolor[rgb]{ .663,  .816,  .557}95.49 & \cellcolor[rgb]{ .663,  .816,  .557}96.71 & 81.57 [RF] \\
    \hline
    \textbf{HT Sensor} & 68.00 & \cellcolor[rgb]{ .663,  .816,  .557}78.00 & \cellcolor[rgb]{ .663,  .816,  .557}72.00 & \cellcolor[rgb]{ .663,  .816,  .557}80.00 & 72.00 [DTW] \\
    \hline
    \textbf{JapaneseVowels} & \cellcolor[rgb]{ .663,  .816,  .557}99.00 & \cellcolor[rgb]{ .663,  .816,  .557}100.00 & \cellcolor[rgb]{ .663,  .816,  .557}99.00 & \cellcolor[rgb]{ .663,  .816,  .557}99.00 & 98.00 \cite{ARkernel} \\
    \hline
    \textbf{KickvsPunch} & 90.00 & \cellcolor[rgb]{ .663,  .816,  .557}100.00 & 90.00 & \cellcolor[rgb]{ .663,  .816,  .557}100.00 & 100.00 \cite{schafer2017multivariate} \\
    \hline
    \textbf{Libras} & \cellcolor[rgb]{ .663,  .816,  .557}99.00 & \cellcolor[rgb]{ .663,  .816,  .557}97.00 & \cellcolor[rgb]{ .663,  .816,  .557}98.00 & \cellcolor[rgb]{ .663,  .816,  .557}97.00 & 95.00 \cite{ARkernel} \\
    \hline
    \textbf{LP1} & 74.00 & 86.00 & 80.00 & 82.00 & 96.00 \cite{schafer2017multivariate} \\
    \hline
    \textbf{LP2} & \cellcolor[rgb]{ .663,  .816,  .557}77.00 & \cellcolor[rgb]{ .663,  .816,  .557}83.00 & \cellcolor[rgb]{ .663,  .816,  .557}80.00 & \cellcolor[rgb]{ .663,  .816,  .557}77.00 & 76.00 \cite{schafer2017multivariate} \\
    \hline
    \textbf{LP3} & 67.00 & \cellcolor[rgb]{ .663,  .816,  .557}80.00 & \cellcolor[rgb]{ .663,  .816,  .557}80.00 & 73.00 & 79.00 \cite{schafer2017multivariate} \\
    \hline
    \textbf{LP4} & 91.00 & 92.00 & 89.00 & 93.00 & 96.00 \cite{schafer2017multivariate} \\
    \hline
    \textbf{LP5} & 61.00 & 66.00 & 62.00 & 67.00 & 71.00 \cite{schafer2017multivariate} \\
    \hline
    \textbf{Movement AAL} & \cellcolor[rgb]{ .663,  .816,  .557}73.25 & \cellcolor[rgb]{ .663,  .816,  .557}79.63 & \cellcolor[rgb]{ .663,  .816,  .557}70.06 & \cellcolor[rgb]{ .663,  .816,  .557}78.34 & 65.61 [SVM-Poly] \\
    \hline
    \textbf{NetFlow} & 94.00 & 95.00 & 93.00 & 95.00 & 98.00 \cite{schafer2017multivariate} \\
    \hline
    \textbf{Occupancy} & \cellcolor[rgb]{ .663,  .816,  .557}71.05 & \cellcolor[rgb]{ .663,  .816,  .557}76.31 & \cellcolor[rgb]{ .663,  .816,  .557}71.05 & \cellcolor[rgb]{ .663,  .816,  .557}72.37 & 67.11 [DTW] \\
    \hline
   {\textbf{OHC}} & \cellcolor[rgb]{ .663,  .816,  .557}99.96 & \cellcolor[rgb]{ .663,  .816,  .557}99.96 & \cellcolor[rgb]{ .663,  .816,  .557}99.96 & \cellcolor[rgb]{ .663,  .816,  .557}99.96 & 99.03 \cite{pei2017multivariate} \\
    \hline
    \textbf{Ozone} & 67.63 & \cellcolor[rgb]{ .663,  .816,  .557}81.50 & \cellcolor[rgb]{ .663,  .816,  .557}79.19 & \cellcolor[rgb]{ .663,  .816,  .557}79.78 & 75.14 [DTW] \\
    \hline
    \textbf{PenDIgits} & \cellcolor[rgb]{ .663,  .816,  .557}97.00 & \cellcolor[rgb]{ .663,  .816,  .557}97.00 & \cellcolor[rgb]{ .663,  .816,  .557}97.00 & \cellcolor[rgb]{ .663,  .816,  .557}97.00 & 95.00 \cite{ARkernel} \\
    \hline
    \textbf{Shapes} & \cellcolor[rgb]{ .663,  .816,  .557}100.00 & \cellcolor[rgb]{ .663,  .816,  .557}100.00 & \cellcolor[rgb]{ .663,  .816,  .557}100.00 & \cellcolor[rgb]{ .663,  .816,  .557}100.00 & 100.00 \cite{schafer2017multivariate} \\
    \hline
    \textbf{UWave} & 97.00 & \cellcolor[rgb]{ .663,  .816,  .557}98.00 & 97.00 & \cellcolor[rgb]{ .663,  .816,  .557}98.00 & 98.00 \cite{schafer2017multivariate} \\
    \hline
    \textbf{Wafer} & \cellcolor[rgb]{ .663,  .816,  .557}99.00 & \cellcolor[rgb]{ .663,  .816,  .557}99.00 & \cellcolor[rgb]{ .663,  .816,  .557}99.00 & \cellcolor[rgb]{ .663,  .816,  .557}99.00 & 99.00 \cite{schafer2017multivariate} \\
    \hline
    \textbf{WalkvsRun} & \cellcolor[rgb]{ .663,  .816,  .557}100.00 & \cellcolor[rgb]{ .663,  .816,  .557}100.00 & \cellcolor[rgb]{ .663,  .816,  .557}100.00 & \cellcolor[rgb]{ .663,  .816,  .557}100.00 & 100.00 \cite{schafer2017multivariate} \\
    \hline
    Arith Mean & 4.63  & 3.29  & 4.17  & 3.17  & - \\
    \hline
    Geo Mean & 3.72  & 2.58  & 3.21  & 2.42  & - \\
    \hline
    Count & 22.00 & 28.00 & 26.00 & 27.00 & - \\
    \hline
    MPCE  & 5.01  & 4.21  & 4.93  & 4.19  & - \\
    \hline

    \end{tabularx}
    \end{adjustbox}
  \label{tab:1}%

\end{table*}%

MLSTM-FCN and MALSTM-FCN is applied on all 35 datasets. We compare our results to the existing reported state-of-the-art models(HULM \cite{pei2017multivariate}, HCRF \cite{quattoni2007hidden}, NL \cite{jaakkola2000discriminative}, FKL \cite{jaakkola2000discriminative}, ARKernel \cite{ARkernel}, LPS \cite{LPS}, mv-ARF \cite{mvARF}, SMTS \cite{SMTS}, WEASEL+MUSE \cite{schafer2017multivariate}, and dUFS \cite{UFS})  of each dataset. Additionally, we compare our models with LSTM-FCN \cite{karim2017lstm}, ALSTM-FCN \cite{karim2017lstm}. Alongside these models, we also obtain baselines for these datasets by testing them on DTW, Random Forest, SVM with a linear kernel, SVM with a 3rd degree polynomial kernel and choose the highest score as the baseline.

Due to the general variance of deep learning algorithms, reproducing exact results is particularly onerous. For replicability, we ran the experiments 3-5 times on various datasets. All the results are similar, where the maximum variance of the accuracy is 3\%. The results presented in Table \ref{tab:1} are obtained when the training loss is a minimum. The weights of the models trained on all of these datasets are provided online. In addition, we provide our training and evaluation scripts that will simplify the replication of similar results. \footnotemark

\footnotetext{The codes and weights of all models are available at \url{https://github.com/houshd/MLSTM-FCN}}

Table \ref{tab:1} compares the performance of various models with MLSTM-FCN and MALSTM-FCN. We define performance as the classification accuracy of a model on a particular dataset. Two datasets, ``Activity'' and ``Action 3d'', required a strided temporal convolution (stride 2) prior to the LSTM branch to reduce the amount of memory consumed when using the MALSTM-FCN model, because the models were too large to fit on a single GTX 1080 Ti processor otherwise. Both of the proposed models, MLSTM-FCN and MALSTM-FCN, outperform the state-of-the-art models (SOTA) on 28 and 27 out of the 35 datasets of this experiment respectively. ``Activity'' is one of the few datasets where the proposed models did not outperform the SOTA model. We postulate that the low performance is due to the large stride of the convolution prior to the LSTM branch, which led to a loss of valuable information.

\begin{figure}[htpb]
\centering
\fbox{
\includegraphics[width=0.8\linewidth]{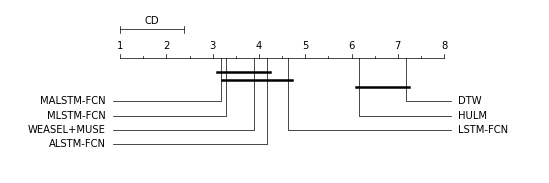}
}
\caption{Critical difference diagram of the arithmetic means of the ranks on all 35 datasets}
\label{fig:critical_diff}

\end{figure}

MLSTM-FCN and MALSTM-FCN have an average arithmetic rank of 3.29 and 3.17 respectively, and a geometric rank of 2.58 and 2.42 respectively. Fig. \ref{fig:critical_diff} depicts the superiority of the proposed models over the top existing models through a critical difference diagram (that applies a Nemenyi test \cite{nemenyi1962distribution}) of the average arithmetic ranks. 

We perform a Wilcoxon signed-rank test to compare all models that were tested on all 35 datasets, as shown in Table \ref{tab:pvaltab}. A  Dunn-Sidak correction \cite{vsidak1967rectangular} is applied to control the familywise error rate, resulting in the adjusted significance of 0.0028. We statistically conclude that the proposed models have a performance score higher than the remaining model as the p-values are below 0.28 percent. The Wilcoxon signed-rank test also demonstrates the performance of MLSTM-FCN and MALSTM-FCN to be the same. Both MLSTM-FCN and MALSTM-FCN perform significantly better than LSTM-FCN and ALSTM-FCN. This indicates the \textit{squeeze-and-excitation} block enhances performance significantly on multivariate time series classification through modeling the inter-dependencies between the variables.

 
\def\tabularxcolumn#1{m{#1}}

\newcolumntype{q}{>{\centering \arraybackslash\hsize=1.15\hsize}X}
\newcolumntype{r}{>{\centering \arraybackslash \hsize=0.9\hsize}X}

\begin{table*}[htpb]
\scriptsize
\centering
\caption{Wilcoxon signed-rank test comparison of Each Model. Red cells denote models where we fail to reject the hypothesis and claim that the models have similar performance.}
\label{tab:pvaltab}

\begin{adjustbox}{width=1 \textwidth, totalheight = 2.3in}

\begin{tabularx}{1.8 \textwidth}{|q|r|q|r|q|r|r|r|r|r|r|r|r|r|r|r|r|r|}
    \hline
          & LSTM-FCN & MLSTM-FCN & ALSTM-FCN & MALSTM-FCN & DTW   & SVM Lin.   & SVM Poly  & RF    & NL    & FKL   & HCRF  & HULM  & dUFS  & SMTS  & LPS   & mv-ARF & ARKernel \\
    \hline
    MLSTM-FCN & 2.76E-04 &       &       &       &       &       &       &       &       &       &       &       &       &       &       &       &  \\
    \hline
    ALSTM-FCN & \cellcolor[rgb]{ 1,  .78,  .808}\textcolor[rgb]{ .612,  0,  .024}{2.41E-01} & 1.93E-03 &       &       &       &       &       &       &       &       &       &       &       &       &       &       &  \\
    \hline
    MALSTM-FCN & 4.40E-04 & \cellcolor[rgb]{ 1,  .78,  .808}\textcolor[rgb]{ .612,  0,  .024}{2.48E-01} & 3.76E-04 &       &       &       &       &       &       &       &       &       &       &       &       &       &  \\
    \hline
    DTW   & 8.22E-08 & 4.72E-09 & 6.98E-08 & 3.67E-09 &       &       &       &       &       &       &       &       &       &       &       &       &  \\
    \hline
    SVM Lin.   & 1.11E-10 & 1.11E-10 & 1.31E-10 & 1.11E-10 & 1.31E-10 &       &       &       &       &       &       &       &       &       &       &       &  \\
    \hline
    SVM Poly  & 1.54E-10 & 1.11E-10 & 1.54E-10 & 1.11E-10 & 2.14E-10 & 2.81E-10 &       &       &       &       &       &       &       &       &       &       &  \\
    \hline
    RF    & 3.68E-10 & 1.54E-10 & 4.56E-10 & 2.26E-10 & 1.88E-09 & 1.17E-10 & 1.54E-10 &       &       &       &       &       &       &       &       &       &  \\
    \hline
    NL    & 1.11E-10 & 1.11E-10 & 1.11E-10 & 1.11E-10 & 1.24E-10 & 1.17E-10 & 1.46E-10 & 1.82E-10 &       &       &       &       &       &       &       &       &  \\
    \hline
    FKL   & 1.11E-10 & 1.31E-10 & 1.24E-10 & 1.24E-10 & 1.63E-10 & 1.11E-10 & 1.17E-10 & 1.17E-10 & 1.31E-10 &       &       &       &       &       &       &       &  \\
    \hline
    HCRF  & 1.11E-10 & 1.11E-10 & 1.11E-10 & 1.11E-10 & 1.72E-10 & 1.11E-10 & 1.11E-10 & 1.31E-10 & 1.31E-10 & 1.63E-10 &       &       &       &       &       &       &  \\
    \hline
    HULM  & 1.17E-10 & 1.38E-10 & 1.31E-10 & 1.11E-10 & 1.63E-10 & 1.11E-10 & 1.11E-10 & 1.11E-10 & 1.11E-10 & 1.46E-10 & 1.46E-10 &       &       &       &       &       &  \\
    \hline
    dUFS  & 1.06E-09 & 2.09E-09 & 2.20E-09 & 2.09E-09 & 1.17E-10 & 1.11E-10 & 1.11E-10 & 1.11E-10 & 1.11E-10 & 1.11E-10 & 1.11E-10 & 1.11E-10 &       &       &       &       &  \\
    \hline
    SMTS  & 5.22E-09 & 1.05E-08 & 1.05E-08 & 7.42E-09 & 4.32E-10 & 1.11E-10 & 1.11E-10 & 1.11E-10 & 1.11E-10 & 1.11E-10 & 1.11E-10 & 1.11E-10 & 5.35E-10 &       &       &       &  \\
    \hline
    LPS   & 1.41E-08 & 1.34E-08 & 1.80E-08 & 8.19E-09 & 4.94E-10 & 1.11E-10 & 1.11E-10 & 1.11E-10 & 1.11E-10 & 1.11E-10 & 1.11E-10 & 1.11E-10 & 4.09E-10 & 1.16E-08 &       &       &  \\
    \hline
    mv-ARF & 1.10E-08 & 4.27E-09 & 6.39E-09 & 2.44E-09 & 4.56E-10 & 1.11E-10 & 1.11E-10 & 1.11E-10 & 1.11E-10 & 1.11E-10 & 1.11E-10 & 1.11E-10 & 2.88E-10 & 7.60E-09 & 7.80E-09 &       &  \\
    \hline
    ARKernel & 4.27E-09 & 1.61E-09 & 1.61E-09 & 1.38E-09 & 5.79E-10 & 1.11E-10 & 1.11E-10 & 1.11E-10 & 1.11E-10 & 1.11E-10 & 1.11E-10 & 1.11E-10 & 5.64E-10 & 1.28E-08 & 1.05E-08 & 9.05E-09 &  \\
    \hline
    WEASEL MUSE & 2.84E-09 & 1.05E-08 & 6.71E-09 & 9.51E-09 & 1.31E-10 & 1.11E-10 & 1.11E-10 & 1.11E-10 & 1.11E-10 & 1.11E-10 & 1.11E-10 & 1.11E-10 & 2.32E-09 & 2.44E-09 & 1.79E-09 & 1.18E-09 & 5.07E-10 \\
    \hline

\end{tabularx}
\end{adjustbox}

\end{table*}

\subsection{Ablation Tests}

\label{ablation}

An ablation study is conducted to determine the effect of dimension shuffle on the input to the LSTM block of the proposed models. We compare the MLSTM-FCN with and without dimension shuffle on all 35 datasets, keeping the number of LSTM cells the same as obtained via grid search for the original models. All other parameters are kept constant. MLSTM-FCN without dimension shuffle took approximately 32 hours to process all the datasets on a GTX 1080 Ti GPU. In comparison, MLSTM-FCN with dimension shuffle required 13 hours to process all the datasets. 

 \newcolumntype{C}{>{\centering\arraybackslash}X}
 
 \newcolumntype{L}{@{}>{\iffalse}l<{\fi}}



 \begin{table*}[htpb]
 \centering
 \caption{Comparison of MLSTM-FCN With and Without Dimension Shuffle }
\label{tab:perf_tab}
\begin{adjustbox}{width=0.6 \linewidth}

 \begin{tabularx}{1 \textwidth}{|C|C|C|C|}
\hline
         & {MLSTM-FCN With Dimension Shuffle} & {MLSTM-FCN Without Dimension Shuffle} \\
    \hline
    MPCE & {4.21} & 4.86 \\
    \hline
    Time (hrs) & 13   & 32 \\
    \hline

    \end{tabularx}
    \end{adjustbox}
  \label{tab:ablation}%

\end{table*}%

The purpose of this study is to determine the impact of the dimension shuffle operation on classification accuracy. Due to the dimension shuffle operation, the time required for training and evaluation of models is significantly reduced in several cases where the number of variables is less than the number of time steps. A Wilcoxon signed-rank test obtains a p-value of 0.136, indicating that we cannot successfully reject the null-hypothesis of the test. This demonstrates the performance of a model when the dimension shuffle operation is applied is statistically the same as when not applied. MLSTM-FCN with dimension shuffle has an MPCE of 4.21. In contrast, an MLSTM-FCN without dimension shuffle obtained a higher MPCE of 4.86. Table \ref{tab:ablation} summarizes how dimension shuffle affects MLSTM-FCN. In other words, the dimension shuffle operation reduces the processing time by 59 percent while maintaining the same classification accuracy.

\section{Conclusion \& Future Work}
\label{conclusion}
The two proposed models attain state-of-the-art results in most of the datasets tested, 28 out of 35 datasets. Each of the proposed models requires minimal preprocessing and feature extraction. Furthermore, the addition of the \textit{squeeze-and-excitation} block improves the performance of LSTM-FCN and ALSTM-FCN significantly. We provide a comparison of our proposed models to other existing state-of-the-art algorithms. 

The proposed models will be beneficial in various multivariate time series classification tasks, such as activity recognition, or action recognition. The proposed models can quickly be deployed in real-time systems and embedded systems because the proposed models are small and efficient. Further research is being done to better understand why the \textit{squeeze-and-excitation} block does not match the performance of the general LSTM-FCN or ALSTM-FCN models on a couple of datasets. 

\pagebreak
\appendix
\section{Variable Definitions}
 \newcolumntype{C}{>{\centering\arraybackslash}X}

 \newcolumntype{L}{@{}>{\iffalse}l<{\fi}}



 \begin{table*}[htpb]
 \centering
 \caption{Definition of all variables}
\label{tab:perf_tab}

\begin{adjustbox}{width=0.71 \linewidth}

 \begin{tabularx}{1.55 \textwidth}{|c|C|c|}

       \hline
    Variable & Definition & {First Introduced} \\
    \hline
    $\mathbf h_t$ & Hidden vector at time step $t$ & 2.1 \\
    \hline
    $\mathbf I$   & Projection matrix & 2.1 \\
    \hline
    $l$   & Layer & 2.1 \\
    \hline
    $\sigma$ & Sigmoid function & 2.1 \\
    \hline
    $t$   & Time step & 2.1 \\
    \hline
    $tanh$ & Hyperbolic tangent function & 2.1 \\
    \hline
    $\mathbf W$   & Weight matrix & 2.1 \\
    \hline
    $\mathbf x_t$ & Input vector at time step $t$ & 2.1 \\
    \hline
    $\mathbf y_t$ & Prediction at time step $t$ & 2.1 \\
    \hline
    $\odot$ & Elementwise multiplication & 2.2 \\
    \hline
    $\mathbf c$   & Cell gate & 2.2 \\
    \hline
    $\mathbf f$   & Forget gate & 2.2 \\
    \hline
    $\mathbf g$   & Activation function & 2.2 \\
    \hline
    $\mathbf h$   & Hidden vector & 2.2 \\
    \hline
    $\mathbf m$   & Memory vector & 2.2 \\
    \hline
    $\mathbf o$   & Output gate & 2.2 \\
    \hline
    $\mathbf u$   & Input gate & 2.2 \\
    \hline
    $a$   & Alignment & 2.3 \\
    \hline
    $\alpha_{ij}$ & Weight & 2.3 \\
    \hline
    $b_i$ & Annotation & 2.3 \\
    \hline
    $e_{ij}$ & Energy of element & 2.3 \\
    \hline
    $i$   & Output position & 2.3 \\
    \hline
    $j$   & Input position & 2.3 \\
    \hline
    $T_x$ & Maximum length of input sequence x & 2.3 \\
    \hline
    $V$   & context vector & 2.3 \\
    \hline
    $v_i$ & context vector & 2.3 \\
    \hline
    $\nu_{}$ & RNN hidden state & 2.3 \\
    \hline
    $*$     & Convolution operation & 2.4 \\
    \hline
    $\mathbf F_{sq}(\mathbf u_c)$ & Channel-wise multiplication between the feature map and the scale & 2.4 \\
    \hline
    $\textbf{F}_{tr}$ & Computational unit for any transformation & 2.4 \\
    \hline
    $\mathbf F_{ex}$ & Parameterized as a neural network & 2.4 \\
    \hline
    $\textbf{s}_c)$ & Channel-wise global average over the temporal dimension $T$ & 2.4 \\
    \hline
    $H$   & Spatial dimension & 2.4 \\
    \hline
    $r$   & Reduction ratio & 2.4 \\
    \hline
    $T$   & Temporal dimension & 2.4 \\
    \hline
    \textbf{U}   & Outputs of $\textbf{F}_{tr}$ & 2.4 \\
    \hline
    $\mathbf v_c^s$ & 2D spatial kernel on channel $c$ & 2.4 \\
    \hline
    $W$   & Spatial dimension & 2.4 \\
    \hline
    $\textbf{W}_1$ & Learnable parameters of $\mathbf F_{ex}$ & 2.4 \\
    \hline
    $\textbf{W}_2$ & Learnable parameters of $\mathbf F_{ex}$ & 2.4 \\
    \hline
    $\textbf{X}$   & Image of shape $H$x$W$x$C$ & 2.4 \\
    \hline
    $\widetilde{\textbf{x}}_c$ & Output of the block rescaled & 2.4 \\
    \hline
    \textbf{z}   & Channel wise statistic & 2.4 \\
    \hline
    $\textbf{z}_c$ & $c$th elment of \textbf{z} & 2.4 \\
    \hline
    $\delta$ & ReLU activation function & 2.4 \\
    \hline
    $\sigma$     & Sigmoid activation function & 2.4 \\
    \hline
    $G_s$ & Number of output feature maps for stage $s$ & 3.1 \\
    \hline
    $M$   & Number of variables processed per time step & 3.1 \\
    \hline
    $P$   & Total number of additional parameters & 3.1 \\
    \hline
    $Q$   & Maximum number of time steps amongst all variables & 3.1 \\
    \hline
    $R_s$ & Repeated block number for stage s & 3.1 \\
    \hline
    $S$   & Number of stages & 3.1 \\
    \hline
    $s$   & Stage & 3.1 \\
    \hline
    $C_i$ & Contribution of each class & 4 \\
    \hline
    $d$   & Number of input units to the weight tensor & 4 \\
    \hline
    $Gw_i$ & loss scaling weight for the i-th class & 4 \\
    \hline
    $N$   & number of samples in the dataset & 4 \\
    \hline
    $N_{C_i}$ & number of samples that belong to class $C_i$ & 4 \\
    \hline
    $U$   & uniform distribution & 4 \\
    \hline
    $K$   & dataset & 4.1 \\
    \hline
    $MPCE$ & mean per class error & 4.1 \\
    \hline
    $\textbf{N}$   & number of datasets & 4.1 \\
    \hline
    $PCE_k$ & per class error for dataset k & 4.1 \\
    \hline

    \end{tabularx}
    \end{adjustbox}
  \label{appendtab:1}%

\end{table*}%

\bibliographystyle{elsarticle-num}

\bibliography{biblio.bib}

\end{document}